\documentclass[lettersize,journal]{IEEEtran}
\usepackage[switch]{lineno}
\usepackage{amsmath,amsfonts}
\usepackage{threeparttable}
\usepackage{algorithm}
\usepackage{algpseudocode}
\usepackage{array}
\usepackage{textcomp}
\usepackage{stfloats}
\usepackage{url}
\usepackage{verbatim}
\usepackage{graphicx}
\usepackage{cite}
\usepackage[numbers,sort&compress]{natbib}
\usepackage{color}

\usepackage{amssymb}
\makeatletter

\newcommand{\Rmnum}[1]{\expandafter\@slowromancap\romannumeral #1@}

\makeatother
\usepackage{multicol}
\usepackage{multirow}
\usepackage{booktabs}
\usepackage{bbding}
\usepackage{longtable}
\usepackage{supertabular}
\usepackage{pifont}
\usepackage{float}
\usepackage{enumitem}
\usepackage{afterpage}

\usepackage[labelformat=simple]{subcaption}
\begin{document}

\title{Cooperative Autonomous Driving in Diverse Behavioral Traffic: A Heterogeneous Graph Reinforcement Learning Approach}


\author{Qi Liu, Xueyuan Li, Zirui Li, Juhui Gim$^{*}$
\thanks{Acknowledgement: This research is funded by Global - Learning \& Academic research institution for Master’s·PhD students, and Postdocs (LAMP) Program of the National Research Foundation of Korea (NRF) grant funded by the Ministry of Education (No. RS2024-00444460) and "Regional Innovation Strategy (RIS)" through the National Research Foundation of Korea (NRF) funded by the Ministry of Education (MOE)(2021RIS-003).}
\thanks{$^{*}$Corresponding authors: Juhui Gim}
\thanks{Qi Liu, Xueyuan Li, and Zirui Li are with the School of Mechanical Engineering, Beijing Institute of Technology, Beijing, China. (E-mails: 3120195257@bit.edu.cn;  lixueyuan@bit.edu.cn; 3120195255@bit.edu.cn.)}
\thanks{Juhui Gim is with the School of Mechatronics Engineering, Changwon National University, Gyeongsangnam-do, South Korea. Email: juhuigim@changwon.ac.kr.}
}


\maketitle

\begin{abstract}
Navigating heterogeneous traffic environments with diverse driving styles poses a significant challenge for autonomous vehicles (AVs) due to their inherent complexity and dynamic interactions. This paper addresses this challenge by proposing a heterogeneous graph reinforcement learning (GRL) framework enhanced with an expert system to improve AV decision-making performance. Initially, a heterogeneous graph representation is introduced to capture the intricate interactions among vehicles. Then, a heterogeneous graph neural network with an expert model (HGNN-EM) is proposed to effectively encode diverse vehicle features and produce driving instructions informed by domain-specific knowledge. Moreover, the double deep Q-learning (DDQN) algorithm is utilized to train the decision-making model. A case study on a typical four-way intersection, involving various driving styles of human vehicles (HVs), demonstrates that the proposed method has superior performance over several baselines regarding safety, efficiency, stability, and convergence rate, all while maintaining favorable real-time performance.

\end{abstract}

\begin{IEEEkeywords}
Autonomous vehicle, decision-making, graph reinforcement learning, heterogeneous graph neural network, heterogeneous traffic.
\end{IEEEkeywords}

\section{Introduction}
\begin{figure}[t]
  \centering
  \includegraphics[scale=0.90]{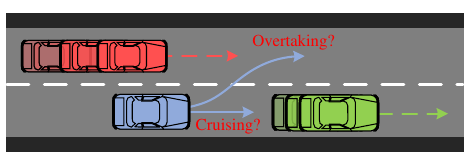}
  \caption{A schematic of scenario with different styles of HVs. The ego AV (blue) must balance safety and efficiency, while comprehensively considering interactions with the aggressive HV (red) and conservative HV (green).}
  \vspace{-10pt}
  \label{scenario_example}
\end{figure}

\IEEEPARstart{T}{he} decision-making for autonomous vehicles (AVs) in modern traffic is highly challenging due to the intricate interactions with these heterogeneous traffic participants. Among these participants, human-driven vehicles (HVs) present significant variability in traits and intentions \cite{9849019}. Considering the Fig. \ref{scenario_example} as an intuitive example. Aggressive HVs often exhibit higher speeds and risk-prone behaviors, showing little regard for other drivers during interactions, whereas conservative HVs prioritize safety, maintaining slower speeds and cautious maneuvers. Normal HVs, in contrast, display a driving style that falls between these two extremes. AVs must carefully negotiate with different styles of HVs to execute reasonable driving behaviors, achieving a superior balance between driving safety and efficiency. 

A comprehensive overview of general decision-making techniques for AVs is presented in \cite{overview2}. Various studies have addressed the challenges of modeling interactions among heterogeneous traffic participants. In \cite{10529605}, the driving styles of HVs are inferred from prior knowledge, then a level-k cooperative game is proposed to generate driving instructions. In \cite{10588863}, interactions among heterogeneous HVs are directly modeled through the self-attention mechanism based on their motion features, then the actor-critic (AC) algorithm is utilized for policy generation. A similar approach targeting the merging scenario is explored in \cite{10138695}. However, these approaches fail to accurately represent the intricate interactions among diverse HVs, limiting their decision-making performance in such heterogeneous traffic.




For encoding heterogeneous traffic information, graph reinforcement learning (GRL)-based methods have demonstrated significant potential. A notable approach in \cite{chen2021graph} models a highway ramping scenario as an undirected graph, facilitating cooperative lane-change decisions for heterogeneous AVs. Building upon this, a multi-view graph representation technique is introduced in \cite{xu2023multi} to more effectively capture vehicle interactions. In \cite{cai2022dq}, scenarios involving heterogeneous vehicles are represented using both a bird's-eye view (BEV) image and an undirected graph, with the graph attention network (GAT) \cite{velickovic2017graph} encoding the information to generate AV maneuvers. Moreover, in \cite{10530461}, an intersection with pedestrians and vehicles of varying driving styles is constructed. Then, a fusion framework consisting of spatial-temporal graph neural network (STGNN), state inference, and trajectory prediction is proposed for AV's decision-making. However, these studies depend on homogeneous technologies to process the features of heterogeneous traffic, which limits the potential for further improvements in decision-making performance.

In conclusion, current approaches still encounter several limitations:
\begin{itemize}[leftmargin=*]
    \item \textbf{Limitations on heterogeneous interaction modeling:} Existing studies fail to comprehensively and accurately model the intricate interactions among diverse traffic participants, limiting AVs' understanding of heterogeneous traffic scenarios and reducing decision-making performance.
    \item \textbf{Limitations on policy generation:} Most studies still depend on homogeneous techniques for processing heterogeneous features, potentially leading to suboptimal or unsafe decisions and limiting the training efficiency of decision-making models.
\end{itemize}

The objective of this paper is to tackle the above decision-making challenges of AVs' current technologies in such heterogeneous traffic by focusing on two critical aspects: 1) Accurate and comprehensive modeling of interactions among heterogeneous traffic participants. 2) Efficient encoding of heterogeneous traffic information and generation of expert-level driving strategies that can well balance safety and efficiency. Specifically, a heterogeneous graph neural network with an expert model (HGNN-EM) is proposed to satisfy the above demands. The main contributions are summarized as follows:

 \begin{itemize}[leftmargin=*]
    \item A heterogeneous graph representation is presented to comprehensively characterize the mutual effects between the ego AV and various styles of HVs, including driving risk evaluation, acceleration difference, and relative distance.
    \item A HGNN-EM is proposed to effectively encode pertinent features from heterogeneous traffic and generate refined driving instructions. By integrating GRL-based policies with expert model-based policies, this approach significantly enhances the decision-making capabilities of AVs.
    \item Comprehensive experiments are conducted on a typical four-way intersection with heterogeneous traffic participants. Results demonstrate that the proposed method outperforms baseline approaches in terms of safety, efficiency, stability, and convergence speed.
\end{itemize}


\begin{figure*}[thpb]
  \centering
  \captionsetup{justification=justified, singlelinecheck=false}
  \includegraphics[scale=0.302]{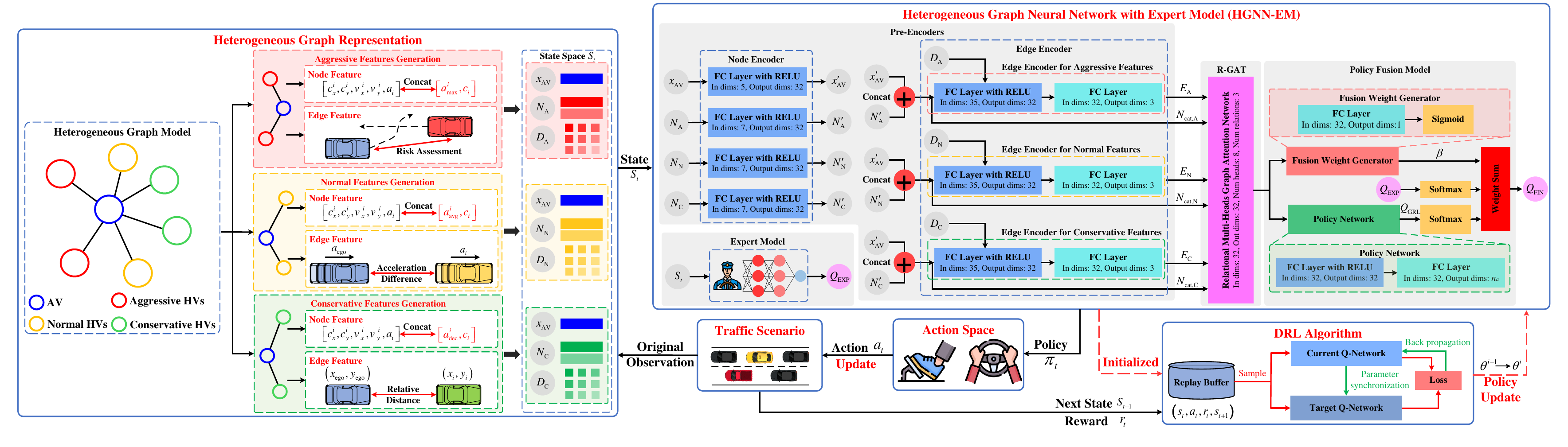}
  \caption{The schematic diagram of the proposed method.}
  \vspace{-10pt}
  \label{Methods}
\end{figure*}

\section{Problem Formulation}
\label{2}

The formulation of the decision-making process within the GRL-based framework can be modeled as a Markov Decision Process (MDP), formulated by tuple \((\mathcal{S},\mathcal{A},P,R,\gamma)\). Specifically, \(P\) is the transition probability, \(\gamma\in(0,1]\) is the discount factor. The other elements are described as follows:

\subsubsection{State Space $\mathcal{S}$}
The intersection is modeled as a heterogeneous undirected graph, where the state space encompasses features of all HVs and the AV within the intersection, generated through the proposed heterogeneous graph representation. The construction process of the state space will be illustrated in Section \ref{3.a}.

\subsubsection{Action Space $\mathcal{A}$}
The action space of the ego AV consists of five high-level discrete driving commands, including \textit{accelerate}, \textit{slow down}, \textit{cruising}, \textit{change lane to the left}, and \textit{change lane to the right}. 

\subsubsection{Reward Function $R$}
A reward function is designed to encourage the AV to turn left safely at the intersection as fast as possible without collisions. Specifically,
\begin{equation}
\label{reward}
    R=r_{\text{goal}}+r_{\text{col}}+r_{\text{vel}}
\end{equation}
where $r_{\text{goal}}$ denotes the task reward, $r_{\text{goal}}=2$ if the ego AV successfully navigates the intersection, otherwise $r_{\text{goal}}=0$. $r_{\text{col}}$ denotes the collision reward to ensure driving safety, $r_{\text{col}}=-2$ if the ego AV collides with HVs, otherwise $r_{\text{col}}=0$. $r_{\text{vel}}$ denotes the dense velocity reward to encourage driving efficiency; specifically, $r_{\text{vel}}=0.04\cdot\text{min}\left\{v_{\text{ego}}/v_{\text{max}},1 \right\}$, where $v_{\text{max}}$ is the maximum permissible longitudinal speed for AV.


\section{Methods}
\label{3}
The proposed method comprises two primary components: the heterogeneous graph representation and the HGNN-EM, as shown in Fig. \ref{Methods}.


\subsection{Heterogeneous Graph Representation}
\label{3.a}
The intersection scenario is modeled as a graph $\mathcal{G}=\{\mathcal{V}, \mathcal{E}\}$, where \(\mathcal{V}=\{\upsilon_{i},i\in\{1,2,...n\} \}\) denotes a set of node attributes (vehicles); \(\mathcal{E}=\{\varepsilon_{ij},i,j\in\{1,2,...n\} \}\) denotes the set of edge attributes (interactions); $n$ represents the total number of all vehicles. Then, the heterogeneous graph representation is utilized to integrate diverse node and edge attributes, reflecting the different styles of vehicles, to construct the graphical state space at each time step $t$. For ease of description, the symbol $t$ is omitted in the following equation. Specifically:

\begin{equation}
\label{state_s}
    G=[x_{\text{AV}},N_{\text{A}},N_{\text{N}},N_{\text{C}},D_{\text{A}},D_{\text{N}},D_{\text{C}}]
\end{equation}
where $x_{\text{AV}}$ denotes the feature vector of the ego AV related to basic motion, including longitudinal coordinate \(c_{x}\), lateral coordinate \(c_{y}\), longitudinal speed \(v_{x}\), lateral speed \(v_{y}\), and acceleration \(a_{x}\), i.e., $x_{\text{AV}}=[c_{x},c_{y},v_{x},v_{y},a_{x}]$. $N_{\text{A}},N_{\text{N}}$, and $N_{\text{C}}$ denote the node feature matrix for aggressive HVs, normal HVs, and conservative HVs, respectively. Similarly, $D_{\text{A}},D_{\text{N}}$, and $D_{\text{C}}$ denotes the adjacency matrix for aggressive HVs, normal HVs, and conservative HVs, respectively.

\subsubsection{Node Feature Matrix}
The node feature matrix encapsulates the driving features of each vehicle. Feature vectors of all HVs include basic motion characteristics $\widetilde{x}=[c_{x},c_{y},v_{x},v_{y},a_{x}]$ and unique features specific to their driving style:

\begin{itemize}[leftmargin=*]
    \item \textbf{Aggressive HVs:} The maximum acceleration \(a_{\text{max}}\) is used to highlight the vehicle's aggressive driving styles, representing its tendency for risk-prone behaviors.
    \item \textbf{Normal HVs:} The average acceleration over $T$ time steps \(a_{\text{avg}}\) is selected to provide a stable view of its typical behavior in traffic. 
    \item \textbf{Conservative HVs:} The maximum deceleration \(a_{\text{dec}}=-4.5\ \text{m/s}^{2}\) is chosen to emphasize the vehicle's avoidance capability, showcasing its cautious driving manner.
\end{itemize}
Let $\delta^{i}\in\{0\ (\text{aggressive}),1\ (\text{normal}),2\ (\text{conservative})\}$ be the category value of the $i_{th}$ HV, the final node features matrix can be formulated as follows:

\vspace{-10pt}
\begin{equation}
\label{node_feature}
\begin{aligned}
    &N_{\text{A}}=[x_{\text{A}}^{1},x_{\text{A}}^{2},...,x_{\text{A}}^{n_{\text{A}}}], x_{\text{A}}^{i}=[\widetilde{x}_{\text{A}}^{i},a_{\text{max}}^{i}, \delta^{i}]\\
    &N_{\text{N}}=[x_{\text{N}}^{1},x_{\text{N}}^{2},...,x_{\text{N}}^{n_{\text{N}}}], x_{\text{N}}^{i}=[\widetilde{x}_{\text{N}}^{i},a_{\text{avg}}^{i}, \delta^{i}]\\
    &N_{\text{C}}=[x_{\text{C}}^{1},x_{\text{C}}^{2},...,x_{\text{C}}^{n_{\text{C}}}], x_{\text{C}}^{i}=[\widetilde{x}_{\text{C}}^{i},a_{\text{dec}}^{i}, \delta^{i}]\\
\end{aligned}
\end{equation}

\subsubsection{Adjacency Matrix}
The adjacency matrix represents the mutual effects among vehicles through specific edge values $e^{ij}$. To account for HVs' varying driving styles, the ego AV needs to focus on different interaction patterns based on each HV's unique characteristics to make informed and adaptive decisions.

\textbf{$\bullet$ Interaction model for Aggressive HVs}

\begin{figure}[t]
  \centering
  \captionsetup{justification=justified, singlelinecheck=false}
  \includegraphics[scale=0.65]{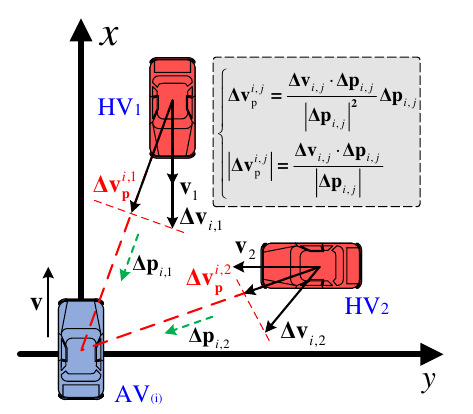}
  \caption{The schematic diagram of the 2D-TTC model.}
  \vspace{-10pt}
  \label{risk}
\end{figure}

Given that aggressive HVs tend to exhibit higher risk-taking behaviors, the ego AV must evaluate the potential for collisions or unsafe situations resulting from their assertive maneuvers. To assess these driving risks, a two-dimensional time to collision (2D-TTC) model \cite{10537107} is established, illustrated in Fig. \ref{risk}.

\noindent Then, the general pattern of the edge value $e_{\text{A}}^{ij}$ within $D_{\text{A}}$ can be formulated as follows:
\begin{equation}
\label{edge_agg}
    e_{\text{A}}^{ij}=\frac{d_{i,j}}{\left|\Delta \mathbf{v}_{\text{p}}^{i, j}\right|}=\frac{d_{i,j}\left|\Delta \mathbf{p}_{i, j}\right|}{\Delta \mathbf{v}_{i, j} \cdot \Delta \mathbf{p}_{i, j}}
\end{equation}
where $d_{i,j}$ is the relative distance between the $i_{th}$ and $j_{th}$ vehicle; $\Delta \mathbf{v}_{\text{p}}^{i, j}$ is the projection of velocity difference vector $\Delta \mathbf{v}_{i, j}$ in the direction of unit projection vector $\Delta \mathbf{p}_{i, j}$. It should be noted that if $e_{\text{A}}^{ij}$ is negative, it will be adjusted to 10 in statistical analysis.

\textbf{$\bullet$ Interaction model for Normal HVs}

Normal HVs tend to exhibit stable driving behavior, making it essential to evaluate the driving similarity between the AV and these HVs to ensure smooth and predictable interactions. Considering that acceleration is an important factor in reflecting the driving styles, the difference of average acceleration over $T$ time steps is chosen to calculate the edge value $e_{\text{N}}^{ij}$ within $D_{\text{N}}$:
\begin{equation}
\label{edge_nor}
    e_{\text{N}}^{ij}=\left|a_{\text{avg}}^{j}-a_{\text{avg}}^{i}\right|
\end{equation}

\textbf{$\bullet$ Interaction model for Conservative HVs}

Conservative HVs generally exhibit cautious driving behavior, the ego AV needs to focus on the potential influences of its actions on these vehicles to adjust its decision-making style dynamically. Given that distance is a direct indicator of driving influence, the relative distance is selected to generate the edge value $e_{\text{C}}^{ij}$ within $D_{\text{C}}$:
\begin{equation}
\label{edge_con}
    e_{\text{C}}^{ij}=\sqrt{\Delta x_{i,j}^{2}+\Delta y_{i,j}^{2}}
\end{equation}
where $\Delta x_{i,j}$ and $\Delta y_{i,j}$ denote the relative longitudinal and lateral distance, respectively.

\subsection{HGNN-EM}
\label{3.2}
The function of the HGNN-EM is to encode the heterogeneous graphical state space efficiently to generate reasonable driving policies. The HGNN-EM consists of four key components: pre-encoders including node and edge encoders, an expert model, a relational graph attention network (R-GAT), and a policy fusion model. 

\subsubsection{Pre-Encoders}
The pre-encoders consist of a node encoder and three edge encoders. The node encoder projects features of different vehicle types—AV, aggressive HVs, normal HVs, and conservative HVs—into a unified embedding space using four separate FC layers with ReLU activation. The embeddings $x_{\text{AV}}^{\prime},N_{\text{A}}^{\prime},N_{\text{N}}^{\prime}$, and $N_{\text{C}}^{\prime}$ for the respective vehicle types are generated for the downstream processing.

The edge encoder uses FC layers to derive learned edge values from node and edge features to offer greater flexibility and adaptability in heterogeneous traffic, enabling the model to more accurately capture intricate mutual effects among vehicles and enhance its representation capability. Let $\Phi_{\text{edge}}$ denote the edge encoder operator, the calculation of the learned edge values can be described as follows:
\begin{equation}
\label{edge_encoder}
\begin{aligned}
    E_{\text{A}}&=\Phi_{\text{edge,A}}\left(N_{\text{cat,A}},D_{\text{A}}\right),N_{\text{cat,A}}=[x_{\text{AV}}^{\prime}\oplus N_{\text{A}}^{\prime}]\\
    E_{\text{N}}&=\Phi_{\text{edge,N}}\left(N_{\text{cat,N}},D_{\text{N}}\right),N_{\text{cat,N}}=[x_{\text{AV}}^{\prime}\oplus N_{\text{N}}^{\prime}]\\
    E_{\text{C}}&=\Phi_{\text{edge,C}}\left(N_{\text{cat,C}},D_{\text{C}}\right),N_{\text{cat,C}}=[x_{\text{AV}}^{\prime}\oplus N_{\text{C}}^{\prime}]\\
\end{aligned}
\end{equation}
where $\oplus$ represents the matrix concatenation operator.

\subsubsection{Expert Model}
The expert model, designed to enhance the GRL-based driving policy with domain-specific guidance, is trained using supervised learning from interactions between expert drivers who can perform highly reasonable driving behaviors and a traffic simulator. The dataset includes the heterogeneous state space (equation (\ref{state_s})) as input and discrete actions $a_{\text{EXP}}\in\mathcal{A}$ as output. Moreover, the expert model shares the same architecture as the HGNN-EM, excluding the fusion weight generator component.


Initially, the output of the dataset is transformed into a one-hot encoded vector $Y=[y_1,y_2,...,y_{n_{\text{a}}}]$, corresponding to the action space, to serve as the target value, where $n_{\text{a}}$ denotes the dimension of the action space. Then, the cross-entropy loss is utilized to ensure that the Q-value table generated by the expert model continuously approximates the expert's action distribution during the training process. The loss function is defined as follows:
\begin{equation}
\label{loss_exp}
    L=-\sum_{i=1}^{n_{\text{a}}} y_i \log \left(\hat{y}_i\right)
\end{equation}
where $\hat{y}_i=\frac{\exp \left(Q_i\right)}{\sum_{j=1}^n \exp \left(Q_j\right)}$ represents the softmax probability of each action, computed from the Q-values predicted by the expert model. The detailed training setting is clarified in Section. \ref{4}.

\subsubsection{R-GAT}
The R-GAT \cite{chen2021r} is the core component of the HGNN-EM, designed to aggregate heterogeneous features information to generate driving policies. The R-GAT extends the conventional graph attention network (GAT) by integrating multiple edge types, which enables more effective processing of heterogeneous graph structures.

First, attention logits $c_{ij}^{k}$ are computed for each relation type $k\in\mathcal{K}=\{0\ (\text{aggressive}),1\ (\text{normal}),2\ (\text{conservative})\}$ based on the corresponding node and edge features:

\begin{equation}
\label{attention}
   c_{ij}^k=\left[\boldsymbol{W}_{\text{E}}^{k}\cdot e_{ij}^k\right] \cdot \left[\boldsymbol{a}^{k}\operatorname{LeakyReLU}\left(\boldsymbol{W}_{\text{N}}^{k} \cdot\left[x_i^k \| x_j^k\right]\right)\right]
\end{equation}
where $\|$ represents the concatenate operator. For the $k_{th}$ relation type, $x_i^k$ and $x_i^k$ denote the feature of the $i_{th}$ and $j_{th}$ node, respectively; $e_{ij}^k$ denote the edge value between the $i_{th}$ and $j_{th}$ node; $\boldsymbol{a}^{k}$ denote the learnable vector; $\boldsymbol{W}_{\text{N}}^{k}$ and $\boldsymbol{W}_{\text{E}}^{k}$ denote the learnable weight matrices for node feature and edge feature, respectively. Then, the attention values are normalized using the softmax function:

\begin{equation}
\label{attention_norm}
    \alpha_{ij}^{k}=\frac{\exp(c_{ij}^{k})}{\sum_{v\in \mathcal{V}_{i}}\exp(c_{iv})}
\end{equation}
Considering all the relation types within the intersection, the feature aggregation process of the R-GAT can be expressed as:

\begin{equation}
\label{feature_cal}
    h_i^{(t)}=\sigma\left(\sum\nolimits_{k\in\mathcal{K}}\sum\nolimits_{v \in \mathcal{V}_i} \alpha_{i v}^{k} \boldsymbol{W}_{\text{N}}^{k} h_v^{(t-1)}\right)
\end{equation}
where $h_i^{(t)}$ denotes the newly generated feature embedding of the $i_{th}$ node, derived from the features of other nodes $h_v^{(t-1)} $ at the preceding time step. $\sigma$ denotes the activation function. Finally, multi-head attention is introduced to further enhance the model learning capacity:

\begin{equation}
\label{feature_final}
    h_i^{(t)}(P)=\|_{p=1}^P h_{i,p}^{(t)}
\end{equation}
where $h_i^{(t),p}$ denotes the feature embedding of the $p_{th}$ head.

\subsubsection{Policy Fusion Model}
The policy fusion model aims to integrate the GRL-based policy $Q_{\text{GRL}}$ with the expert policy $Q_{\text{EXP}}$ to enhance decision-making performance. It comprises two main components: a fusion weight generator and a policy network. 

The fusion weight generator produces a learnable weight value $\beta\in[0,1]$ based on the feature embedding generated by the R-GAT. This mechanism allows the weight values to be dynamically adjusted in response to the feature embeddings of heterogeneous traffic, thereby enhancing the adaptability and flexibility of the fused policy. The policy network further processes the feature embedding from the R-GAT to generate GRL-based policy. The final driving policy can be represented as follows:
\begin{equation}
\label{final_policy}
    Q_{\text{FIN}}=\beta\operatorname{Softmax}(Q_{\text{GRL}})+(1-\beta)\operatorname{Softmax}(Q_{\text{EXP}})
\end{equation}
where $Q_{\text{FIN}}$ denote the final policy. The role of the softmax function is to ensure that the $Q_{\text{GRL}}$ and $Q_{\text{EXP}}$ are scaled consistently before policy fusion. The complete feature extraction procedure of the HGNN-EM is illustrated in Algorithm \ref{HGNN-EM}.

\begin{algorithm}[t]
        \small
        \renewcommand{\algorithmicrequire}{\textbf{Input:}}
	\renewcommand{\algorithmicensure}{\textbf{Output:}}
	\caption{The computing procedure of the HGNN-EM} 
	\label{HGNN-EM} 
	\begin{algorithmic}[1]
		  \Require Original heterogeneous graphical state space $G=[x_{\text{AV}},N_{\text{A}},N_{\text{N}},N_{\text{C}},D_{\text{A}},D_{\text{N}},D_{\text{C}}]$. 
		\Ensure Fused Q-values for AVs $Q_{\text{FIN}}$.
        
            \Statex \hspace{-2em} \textbf{$\bullet$ Pre-processing for the original input}
		\State Calculate the fixed-dimensional node embeddings $x_{\text{AV}}^{\prime},N_{\text{A}}^{\prime},N_{\text{N}}^{\prime}$, and $N_{\text{C}}^{\prime}$ through the node encoder.
            \State Generate the learned edge values by equation (\ref{edge_encoder}).
            
            \Statex \hspace{-2em} \textbf{$\bullet$ Processing of the R-GAT}
            \State Calculate attention logits $c_{ij}^{k}$ for each type of relations by equation (\ref{attention}).
            \State Calculate normalized attention values $\alpha_{ij}^{k}$ for each type of relations by equation (\ref{attention_norm}).
            \State Calculate the aggregated feature embedding $h_{i}^{(t)}$ for each node by equation (\ref{feature_cal}).
            \State Calculate the multi-head feature embedding $h_{i}^{(t)}$ for each node by equation (\ref{feature_final}).
            
            \Statex \hspace{-2em} \textbf{$\bullet$ Processing of the policy fusion model}
            \State Calculate the fusion weight $\beta$ through the fusion weight generator.
            \State Calculate the $Q_{\text{GRL}}$ through the policy network, taking the multi-head feature embedding as input.
            \State Calculate the $Q_{\text{EXP}}$ through the well-trained expert model, taking the heterogeneous graphical state space $G$ as input.
            \State Calculate the final Q values for AV by equation (\ref{final_policy}).
            
            \Statex \hspace{-2em} \textbf{Return:} $Q_{\text{FIN}}$.
	\end{algorithmic} 
    \vspace{-2pt}
\end{algorithm}

\subsection{Policy Optimization}
The Double DQN (DDQN) \cite{van2016deep} algorithm is utilized as the DRL model to train the decision-making model. Two instances of HGNN-EM are initialized: the current Q-network $\theta$ for action value generation, and the target Q-network $\hat{\theta}$ for guiding policy updates. It is worth noting that the parameter of the well-trained expert model is fixed during training.

Initially, a \(T_{\text{r}}\)-step random exploration phase is implemented to broaden the exploration space of AV. Then, AV interacts with the heterogeneous traffic based on the linear decay $\epsilon-{\rm{greedy}}$ strategy. The value of $\epsilon$ at each time step is determined using the following equation:

\begin{equation}
\label{policy}
    \epsilon_{t}=-\frac{(\epsilon_{i}-\epsilon_{f})t}{T_{e}}+\epsilon_{i}
\end{equation}
where \(\epsilon_{i}\) is the initial exploration rate, \(\epsilon_{f}\) is the final exploration rate, and \(T_{e}\) is the max exploration step. Then, the action-selecting strategy can be expressed as:

\begin{equation}
\label{action}
a_{t}=
\left\{
\begin{array}{ll}
\text{random}(\mathcal{A}), & P=\epsilon_{t} \\
\text{argmax}\ Q_{\text{FIN}}(s_{t},a;\theta), & P=1-\epsilon_{t} \\
\end{array}\right.
\end{equation}

\noindent After performing the actions, the interaction trajectory $(s_{t},a_{t},r_{t},s_{t+1})$ is recorded in a replay memory pool $\mathcal{D}$ with capacity $N_{\mathcal{D}}$. During the training phase, a random minibatch \(B\) comprising transitions $(s_{J},a_{J},r_{J},s_{J+1})$ is sampled from $\mathcal{D}$. Subsequently, the training loss can be calculated as follows:

\begin{equation}
\label{loss}
\left\{
\begin{aligned}
    &\hat{y}_{J}=r_{J} + \gamma Q_{\text{FIN}}(s_{J+1},\underset{a}{\textrm{argmax}}\ Q_{\text{FIN}}(s_{J+1},a;\theta);\hat{\theta})\\
    &L(\theta)=(\hat{y}_{J}-Q_{\text{FIN}}(s_{J},a_{J};\theta))^{2}
\end{aligned}\right.
\end{equation}
Finally, the parameters of the two HGNN-EM instances are updated, thereby solving the optimal driving strategies:
\begin{equation}
\label{update}
\left\{
\begin{array}{ll}
\theta\leftarrow \theta-\alpha\nabla_{\theta}L(\theta), & t\ \text{mod}\ T_{c}=0 \\
\hat{\theta}\leftarrow \theta, & t\ \text{mod}\ T_{t}=0 \\
\end{array}\right.
\end{equation}
where $T_{c}$ and $T_{t}$ denote the model update interval of the current Q-network and target Q-network, respectively. $\alpha$ is the learning rate.

\section{Case Study}
\label{4}
The proposed method is evaluated with a case study on a typical four-way intersection. It describes the scenario construction, basic settings, and experimental results. 

\subsection{Scenario Construction}
\begin{figure}[t]
  \centering
  \captionsetup{justification=justified, singlelinecheck=false}
  \includegraphics[scale=0.38]{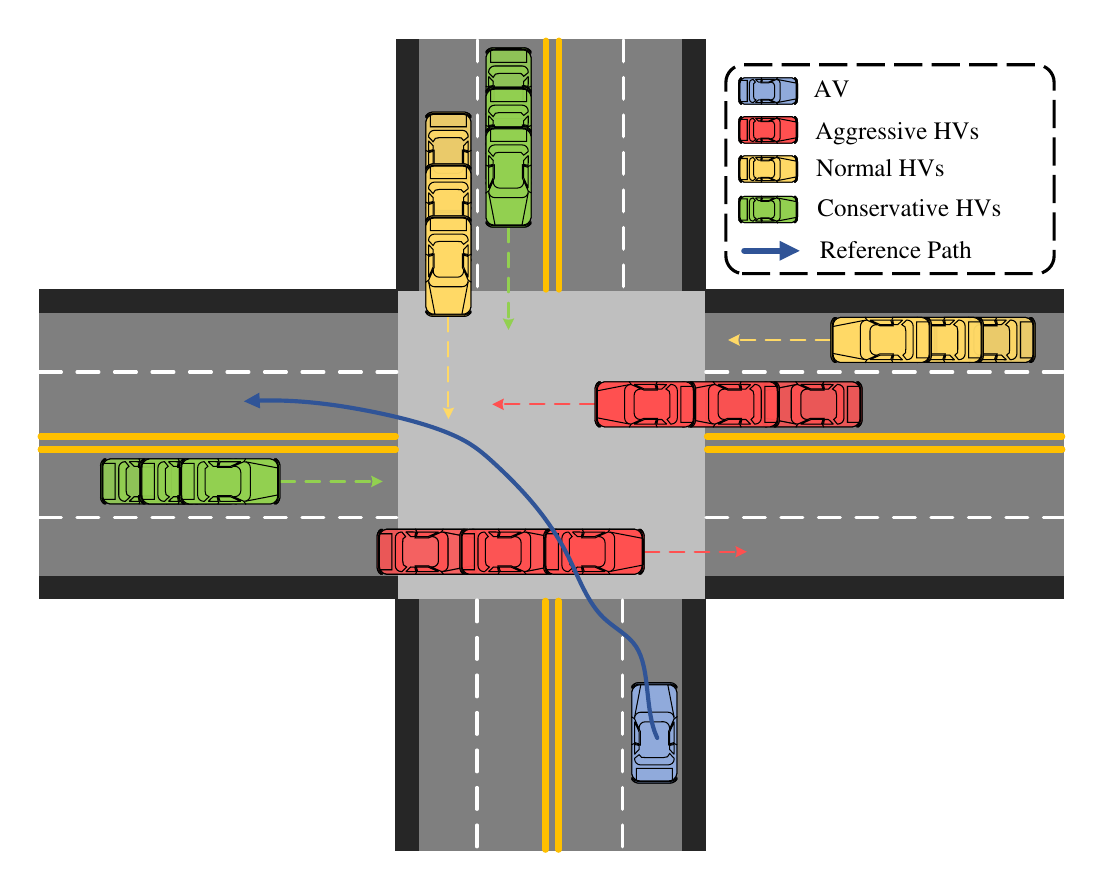}
  \caption{The schematic of the typical four-way intersection.}
  \vspace{-2pt}
  \label{scenario}
\end{figure}

Fig. \ref{scenario} shows a four-way unsigned intersection constructed based on the SUMO \cite{dlr127994} simulator with six HVs, evenly distributed across three driving styles: aggressive, normal, and conservative. HVs travel straight from designated starting areas to the opposite side. The AV's goal is to safely and efficiently navigate the intersection, accounting for varied HV behaviors. Furthermore, a set of key rules is established: 1) AV can obtain information on all HVs through its onboard sensors. 2) AV can directly acquire the driving characteristics of HVs.

The intelligent driver model (IDM) \cite{PhysRevE} and the MOBIL model \cite{kesting2007general} are utilized for longitudinal and lateral controls after decision-making, respectively. Parameters are customized to reflect varied driving styles among HVs, as detailed in Table \ref{para_IDM}. Furthermore, the distinct behavioral patterns of HVs are incorporated to reflect varying levels of interactions at the intersection. Specifically:
\begin{itemize}[leftmargin=*]
    \item The aggressive HVs exhibit higher speed and disregard the right-of-way rules within the intersection, prioritizing speed and assertiveness over safety in conflict situations.
    \item The normal HVs adhere to the right-of-way rules within the intersection, striking a balance between safety and efficiency when conflicts exist. Their behaviors provide a baseline for interaction with the AV and other HVs.
    \item The conservative HVs prioritize safety and respond more cautiously to other vehicles, tending to wait for other vehicles to pass through the intersection first.
\end{itemize}

\begin{table}[t]
\caption{Parameters of IDM for different styles of HVs}
\label{para_IDM}
\begin{center}
\begin{tabular}{ccccc}
\toprule
\multirow{2.2}{*}{\textbf{Symbols}} & \multirow{2.2}{*}{\textbf{Description}} & \multicolumn{3}{c}{\textbf{Value}}\\
\cmidrule{3-5}
~ & ~ & Agg. & Nor. & Con. \\
\midrule 
$v\ (\text{m/s})$ & Longitudinal speed & \multicolumn{3}{c}{Variable}\\
$\Delta v\ (\text{m/s})$ & Relative speed & \multicolumn{3}{c}{Variable}\\
$s\ (\text{m})$ & Distance from leader & \multicolumn{3}{c}{Variable}\\
\midrule 

$a_{\max }\ (\text{m/s}^{2})$ & Maximum acceleration & 4.5 & 3.5 & 2.5\\
$\delta$ & Acceleration exponent & 5 & 4 & 4\\
$v^*\ (\text{m/s})$ & Desired speed & 20 & 16 & 12\\
$s_0\ (\text{m})$ & Minimum gap from leader & 1.2 & 1.6 & 2.0\\
$T\ (\text{s})$ & Safety time gap & 1.0 & 1.5 & 2.0\\
$b_{\mathrm{comf}}\ (\text{m/s}^{2})$ & Comfortable deceleration & 2.0 & 2.0 & 2.0\\
\bottomrule
\end{tabular}
\end{center}
\vspace{-10pt}
\end{table}

\subsection{Experiment Setting}

\subsubsection{Implementation Details}
The initial speed of all vehicles is set to $0\ \text{m/s}$. The observation time length $T$ for the average acceleration is set to 5 time steps. The maximum permissible longitudinal speed $v_{\text{max}}$ for AV is set to $20\ \text{m/s}$. The hyper-parameters for model training are listed in Table \ref{hyper-para}.

\subsubsection{Training Details of the Expert Model}
A keyboard interface has been integrated into SUMO to allow expert drivers to maneuver AVs through intersections, generating a dataset of $(G^{i},a_{\text{EXP}}^{i})$ pairs representing heterogeneous states and expert actions. The dataset consists of $5.5\times10^{5}$ samples, with 70\% for training and 30\% for testing.

The hyper-parameters for training the expert model are listed in Table \ref{hyper-para-em}. During testing, the model is tested for 50,000 times. The accuracy of action generation is calculated, achieving a value of 99.23\%, which underscores the effectiveness and reliability of the trained expert model in replicating expert-level decision-making.

\begin{table}[t]
\caption{Hyper-parameters for Model Training}
\label{hyper-para}
\begin{center}
\begin{tabular}{ccc}
\toprule
\textbf{Symbols} & \textbf{Description} & \textbf{Value}\\
\midrule 
— & Simulation time step & 0.1 s\\
\(M\) & Max training episodes & 150 \\
\(N_{\text{test}}\) & Number of testing episodes & 50 \\
\(T_{M}\) & Max time steps for each episode & 300 \\
\(T_{r}\) & Random exploration step & 9000 \\
\(T_{c}\) & Max exploration step & 10000 \\
\(\epsilon_{i}\) & Initial exploration rate & 0.5 \\
\(\epsilon_{f}\) & Final exploration rate & 0.01 \\
$N_{\mathcal{D}}$ & Replay memory pool capacity & $1\times10^{6}$ \\
\(B\) & Minibatch size & 64 \\
\(\alpha\) & Learning rate & 0.0005 \\
\(T_{c}\) & Update interval for current Q-network & 50 \\
\(T_{t}\) & Update interval for target Q-network & 5000 \\
\(\gamma\) & Discount factor & 0.99\\
\bottomrule
\end{tabular}
\end{center}
\vspace{-5pt}
\end{table}

\begin{table}[t]
\caption{Hyper-parameters for Expert Model Training}
\label{hyper-para-em}
\begin{center}
\begin{tabular}{ccc}
\toprule
\textbf{Description} & \textbf{Value}\\
\midrule 
Number of training epochs & 500 \\
Training batch size & 32 \\
Max training batches for each epoch & 100 \\
Optimizer & Adam \\
Learning rate & 0.0005 \\

\bottomrule
\end{tabular}
\end{center}
\vspace{-5pt}
\end{table}

\begin{table}[t]
\begin{center}
\caption{Baselines for Comparative Experiment}
\label{alg_compare}
\scalebox{1}{
\begin{tabular}{cccc}
\toprule
\textbf{Model} & \textbf{Graph Representation} & \textbf{Backbone}\\
\midrule
GCQ \cite{chen2021graph} & Homogeneous graph & GCN \\
HGNN & Heterogeneous graph & HGNN \\
STGCN \cite{10530461,10588390} & Homogeneous graph & GCN, TCN \\
\textbf{Proposed} & Heterogeneous graph & HGNN-EM \\

\bottomrule
\end{tabular}}
\end{center}
\vspace{-10pt}
\end{table}


\subsubsection{Evaluation Metrics}
For each episode, several evaluation metrics are introduced to evaluate the performance of the AV, specifically:
\begin{itemize}[leftmargin=*]
    \item \textbf{Reward:} The reward for each episode is recorded as the effective evaluation of the overall performance of the AV.
    \item \textbf{Collision number:} The count of collisions performed by the ego AV is recorded as a safety assessment.
    \item \textbf{Average speed:} The average speed (m/s) of AV during driving is computed to evaluate driving efficiency.
    \item \textbf{Traveling time:} The cumulative time (s) for the AV to exit the intersection is recorded to evaluate the efficiency of task accomplishment.
\end{itemize}

\noindent For model testing, one metric is additionally established, specifically:

\begin{itemize}[leftmargin=*]
    \item \textbf{Computational efficiency:} The average computing time (ms) per simulation step is measured to assess the computational efficiency. The test device is a laptop with 32GB RAM, i9-10980HK, RTX4070s.
\end{itemize}

\subsection{Experimental Results and Analysis}
\subsubsection{Baselines}
To assess the performance of our proposed method, we compare it with several state-of-the-art baseline models, all calibrated to match the decision-making problem formulation in Fig. \ref{Methods}. Details of these models are summarized in Table \ref{alg_compare}.

\subsubsection{Results and Analysis}
\begin{figure}[t]
    \centering
    \captionsetup{justification=justified, singlelinecheck=false}
        \begin{subfigure}{\columnwidth} 
            \centering
            \captionsetup{justification=centering, singlelinecheck=false}
            \includegraphics[width=0.7\textwidth]{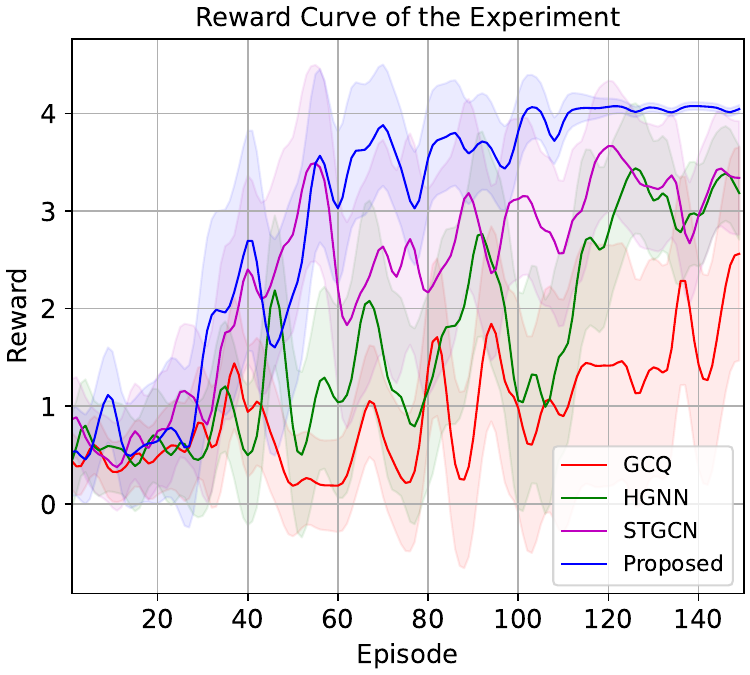}
            \caption{Reward curves of all models.}
            \label{r1_c} 
        \end{subfigure}

        \vspace{1em} 

        \begin{subfigure}{\columnwidth}
            \centering
            \captionsetup{justification=centering, singlelinecheck=false}
            \includegraphics[width=0.85\textwidth]{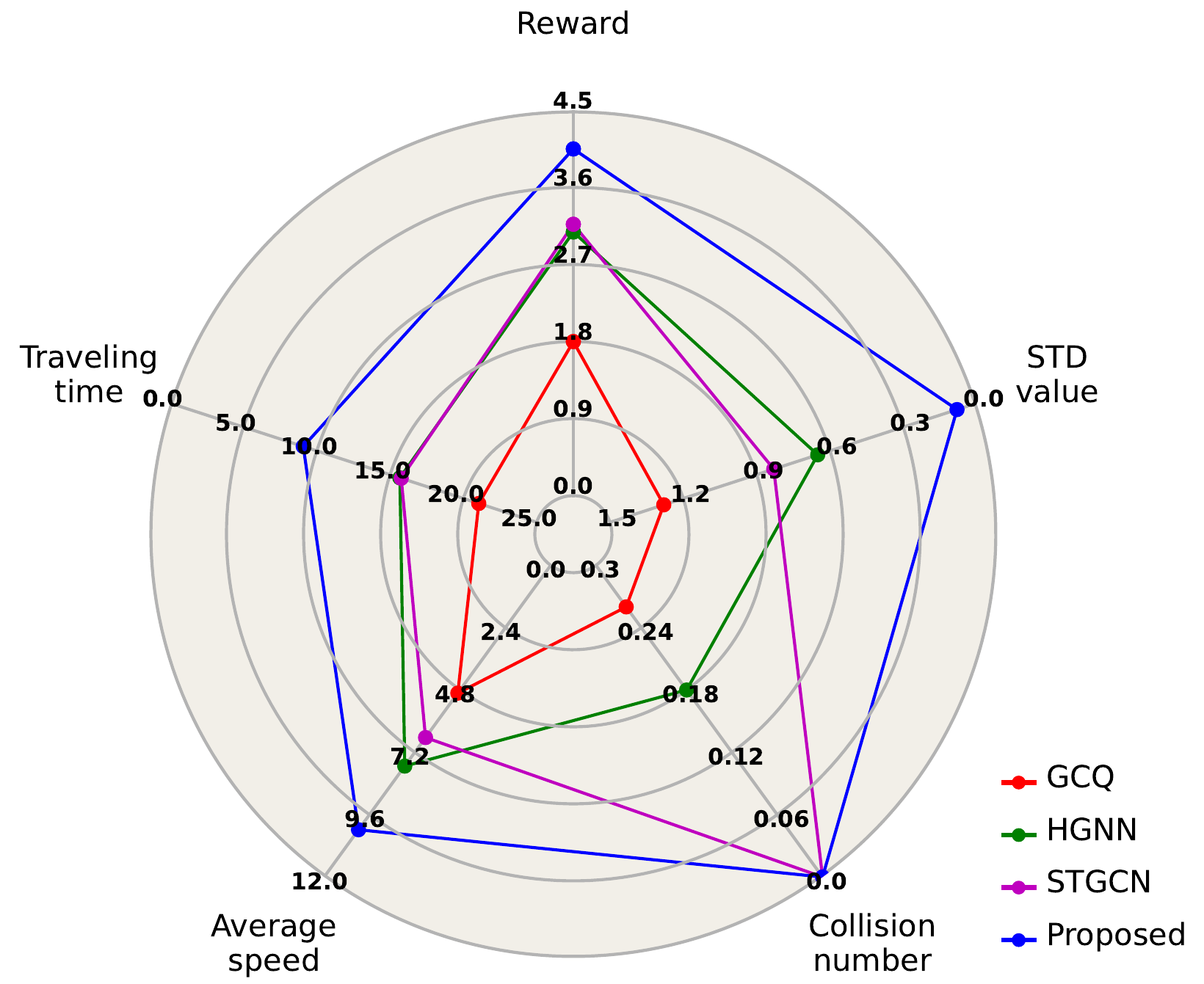}
            \caption{Training results of all models.}
            \label{r2_c} 
        \end{subfigure}

        \vspace{1em}

        \begin{subfigure}{\columnwidth}
            \centering
            \captionsetup{justification=centering, singlelinecheck=false}
            \includegraphics[width=0.85\textwidth]{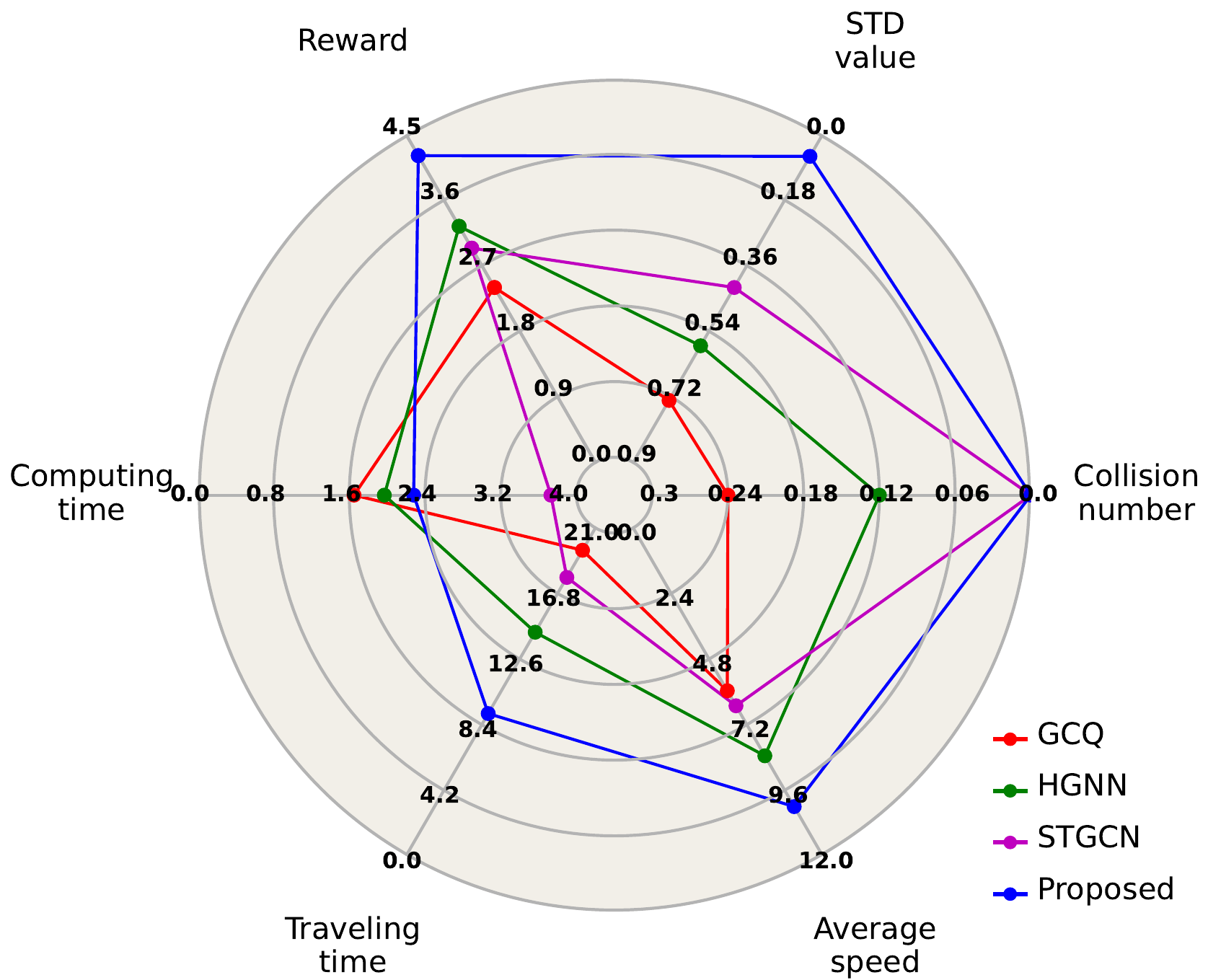}
            \caption{Testing results of all models.}
            \label{r3_c} 
        \end{subfigure}
    \caption{The results of the experiment. The shaded areas in the reward figure show the standard deviation for 3 random seeds.}
    \vspace{-10pt}
    \label{data_com}
\end{figure}

During training, reward curves for three random seeds are plotted in Fig. \ref{r1_c}, with final results averaged after convergence (around 120 episodes). Additionally, the convergent reward trajectories are recorded to compute the standard deviation (STD) to reflect the model's stability. These training results are displayed in Fig. \ref{r2_c}. During testing, each model yields three decision-making instances, and the averaged testing results are summarized in Fig. \ref{r3_c}.

As shown in Fig. \ref{r1_c}, all models finally converge to relatively stable reward intervals, signifying the effectiveness of the training process. As seen in Fig. \ref{r2_c}, the traveling times for all models remain within the max time steps limitation ($T_{M}=300\  \text{steps}=30\ \text{s}$) after achieving convergence. This demonstrates that all models are capable of successfully navigating the intersection. However, the GCQ model exhibits inferior performance across all evaluation metrics, underscoring the limitations of relying solely on graphic techniques without additional enhancements in heterogeneous traffic scenarios.

The HGNN model exhibits superior overall performance compared to the GCN model, particularly evident in the lower STD value. This emphasizes that the incorporation of heterogeneous technologies into the graph representation and the GNN significantly optimizes the decision-making performance and model adaptability in heterogeneous traffic. Nonetheless, there is room for improvement in the driving safety aspect of the HGNN model. Compared to the HGNN model, the STGCN model performs better in STD value and driving safety (zero collisions). This highlights the importance of incorporating temporal networks, facilitating more consistent and stable behaviors, thereby enhancing driving safety and model stability in heterogeneous traffic. 

Notably, the proposed method achieves zero collisions and outperforms other models in terms of reward, average speed, and traveling time, showcasing its strong ability to generate safe and efficient driving strategies in heterogeneous traffic. The lower STD value further underscores its adaptability to the varied driving styles of HVs in heterogeneous traffic.

As seen in Fig. \ref{r3_c}, the testing results align closely with the converged training outcomes depicted in Fig. \ref{r2_c}, further validating the effectiveness of the model training process. The STGCN model exhibits a noticeable increase in computing time, highlighting that the integration of spatial-temporal techniques significantly elevates data processing demands, thereby reducing computational efficiency. In contrast, the computing time of the proposed method remains comparable to that of the GCQ and HGNN models while delivering superior overall decision-making performance. This emphasizes that the proposed method effectively balances performance gains with efficiency consideration.

\section{Conclusion}
\label{5}

This paper introduces a combined approach of heterogeneous GRL and an expert system to enhance AVs' decision-making in heterogeneous traffic. Specifically, a heterogeneous graph representation is proposed to capture intricate interactions, while an HGNN-EM is established to encode the heterogeneous information and generate expert-level driving behaviors. Experiments on a typical four-way intersection demonstrate the superior performance of the proposed method. Future work will focus on predicting the driving styles of HVs and validating our approach across a broader range of traffic scenarios with diverse settings, thereby further enhancing the adaptability and robustness of the proposed method.

\footnotesize
\bibliographystyle{IEEEtran}
\bibliography{myref}

\begin{thebibliography}{10}
\providecommand{\url}[1]{#1}
\csname url@samestyle\endcsname
\providecommand{\newblock}{\relax}
\providecommand{\bibinfo}[2]{#2}
\providecommand{\BIBentrySTDinterwordspacing}{\spaceskip=0pt\relax}
\providecommand{\BIBentryALTinterwordstretchfactor}{4}
\providecommand{\BIBentryALTinterwordspacing}{\spaceskip=\fontdimen2\font plus
\BIBentryALTinterwordstretchfactor\fontdimen3\font minus \fontdimen4\font\relax}
\providecommand{\BIBforeignlanguage}[2]{{%
\expandafter\ifx\csname l@#1\endcsname\relax
\typeout{** WARNING: IEEEtran.bst: No hyphenation pattern has been}%
\typeout{** loaded for the language `#1'. Using the pattern for}%
\typeout{** the default language instead.}%
\else
\language=\csname l@#1\endcsname
\fi
#2}}
\providecommand{\BIBdecl}{\relax}
\BIBdecl

\bibitem{9849019}
S.~Bae, D.~Isele, A.~Nakhaei, P.~Xu, A.~M. Añon, C.~Choi, K.~Fujimura, and S.~Moura, ``Lane-change in dense traffic with model predictive control and neural networks,'' \emph{IEEE Transactions on Control Systems Technology}, vol.~31, no.~2, pp. 646--659, 2023.

\bibitem{overview2}
Q.~Liu, X.~Li, S.~Yuan, and Z.~Li, ``Decision-making technology for autonomous vehicles: Learning-based methods, applications and future outlook,'' in \emph{2021 IEEE International Intelligent Transportation Systems Conference (ITSC)}, 2021, pp. 30--37.

\bibitem{10529605}
S.~Fang, P.~Hang, C.~Wei, Y.~Xing, and J.~Sun, ``Cooperative driving of connected autonomous vehicles in heterogeneous mixed traffic: A game theoretic approach,'' \emph{IEEE Transactions on Intelligent Vehicles}, pp. 1--15, 2024.

\bibitem{10588863}
K.~Chen, B.~Li, R.~Zhang, and X.~Cheng, ``Autonomous intersection management with heterogeneous vehicles: A multi-agent reinforcement learning approach,'' in \emph{2024 IEEE Intelligent Vehicles Symposium (IV)}, 2024, pp. 2255--2260.

\bibitem{10138695}
J.~Liu, W.~Zhao, C.~Wang, C.~Xu, L.~Li, Q.~Chen, and Y.~Lian, ``Eco-friendly on-ramp merging strategy for connected and automated vehicles in heterogeneous traffic,'' \emph{IEEE Transactions on Vehicular Technology}, vol.~72, no.~11, pp. 13\,888--13\,900, 2023.

\bibitem{chen2021graph}
S.~Chen, J.~Dong, P.~Ha, Y.~Li, and S.~Labi, ``Graph neural network and reinforcement learning for multi-agent cooperative control of connected autonomous vehicles,'' \emph{Computer-Aided Civil and Infrastructure Engineering}, vol.~36, no.~7, pp. 838--857, 2021.

\bibitem{xu2023multi}
D.~Xu, P.~Liu, H.~Li, H.~Guo, Z.~Xie, and Q.~Xuan, ``Multi-view graph convolution network reinforcement learning for cavs cooperative control in highway mixed traffic,'' \emph{IEEE Transactions on Intelligent Vehicles}, vol.~9, no.~1, pp. 2588--2599, 2024.

\bibitem{cai2022dq}
P.~Cai, H.~Wang, Y.~Sun, and M.~Liu, ``Dq-gat: Towards safe and efficient autonomous driving with deep q-learning and graph attention networks,'' \emph{IEEE Transactions on Intelligent Transportation Systems}, 2022.

\bibitem{velickovic2017graph}
P.~Velickovic, G.~Cucurull, A.~Casanova, A.~Romero, P.~Lio, and Y.~Bengio, ``Graph attention networks,'' \emph{stat}, vol. 1050, no.~20, pp. 10--48\,550, 2017.

\bibitem{10530461}
J.~Li, D.~Isele, K.~Lee, J.~Park, K.~Fujimura, and M.~J. Kochenderfer, ``Interactive autonomous navigation with internal state inference and interactivity estimation,'' \emph{IEEE Transactions on Robotics}, vol.~40, pp. 2932--2949, 2024.

\bibitem{10537107}
Z.~Li, J.~Gong, Z.~Zhang, C.~Lu, V.~L. Knoop, and M.~Wang, ``Interactive behavior modeling for vulnerable road users with risk-taking styles in urban scenarios: A heterogeneous graph learning approach,'' \emph{IEEE Transactions on Intelligent Transportation Systems}, vol.~25, no.~8, pp. 8538--8555, 2024.

\bibitem{chen2021r}
M.~Chen, Y.~Zhang, X.~Kou, Y.~Li, and Y.~Zhang, ``R-gat: relational graph attention network for multi-relational graphs,'' \emph{arXiv preprint arXiv:2109.05922}, 2021.

\bibitem{van2016deep}
H.~Van~Hasselt, A.~Guez, and D.~Silver, ``Deep reinforcement learning with double q-learning,'' in \emph{Proceedings of the AAAI conference on artificial intelligence}, vol.~30, no.~1, 2016.

\bibitem{dlr127994}
P.~A. Lopez, M.~Behrisch, L.~Bieker-Walz, J.~Erdmann, Y.-P. Fl{\"o}tter{\"o}d, R.~Hilbrich, L.~L{\"u}cken, J.~Rummel, P.~Wagner, and E.~Wie{\ss}ner, ``Microscopic traffic simulation using sumo,'' in \emph{The 21st IEEE International Conference on Intelligent Transportation Systems}, 2018, pp. 2575--2582.

\bibitem{PhysRevE}
\BIBentryALTinterwordspacing
M.~Treiber, A.~Hennecke, and D.~Helbing, ``Congested traffic states in empirical observations and microscopic simulations,'' \emph{Phys. Rev. E}, vol.~62, pp. 1805--1824, Aug 2000. [Online]. Available: \url{https://link.aps.org/doi/10.1103/PhysRevE.62.1805}
\BIBentrySTDinterwordspacing

\bibitem{kesting2007general}
A.~Kesting, M.~Treiber, and D.~Helbing, ``General lane-changing model mobil for car-following models,'' \emph{Transportation Research Record}, vol. 1999, no.~1, pp. 86--94, 2007.

\bibitem{10588390}
Q.~Liu, Y.~Tang, X.~Li, F.~Yang, X.~Gao, and Z.~Li, ``Sif-stgdan: A social interaction force spatial-temporal graph dynamic attention network for decision-making of connected and autonomous vehicles,'' in \emph{2024 IEEE Intelligent Vehicles Symposium (IV)}, 2024, pp. 376--383.

\end{thebibliography}

\vfill

\end{document}